\def\year{2021}\relax
\documentclass[letterpaper]{article} 
\usepackage{aaai21}  
\usepackage{times}  
\usepackage{helvet} 
\usepackage{courier}  
\usepackage[hyphens]{url}  
\usepackage{graphicx} 
\usepackage{natbib}  
\usepackage{caption} 
\usepackage{amsmath,amsfonts,amssymb}
\usepackage{mathtools,nccmath}
\usepackage{varwidth}
\usepackage{hyperref}

\usepackage{booktabs}       

\usepackage{xcolor} 

\usepackage{algorithm}
\usepackage[noend]{algpseudocode}
\usepackage{multirow}

\usepackage[switch]{lineno}

\title{Nested Named Entity Recognition with Partially-Observed TreeCRFs}

\author{
    \Large{\bf{ Yao Fu\textsuperscript{1}\footnote{Equal Contribution.}, Chuanqi Tan\textsuperscript{2}\footnotemark[1]\footnote{Corresponding author.}, Mosha Chen\textsuperscript{2},
    Songfang Huang\textsuperscript{2}, Fei Huang\textsuperscript{2}}} \\
    \Large{\textsuperscript{1}University of Edinburgh \enskip \enskip\enskip\enskip \textsuperscript{2}Alibaba Group}  \\ 
    \Large{yao.fu@ed.ac.uk  \enskip \{chuanqi.tcq; chenmosha.cms; songfang.hsf; f.huang\}@alibaba-inc.com}
}

\begin{document}
\maketitle
	

\begin{abstract}
	Named entity recognition (NER) is a well-studied task in natural language processing. However, the widely-used sequence labeling framework is difficult to detect entities with nested structures.  
	In this work, we view nested NER as constituency parsing with partially-observed trees and model it with partially-observed TreeCRFs.
	Specifically, we view all labeled entity spans as observed nodes in a constituency tree, and other spans as latent nodes. 
	With the TreeCRF we achieve a uniform way to jointly model the observed and the latent nodes. 
	To compute the probability of partial trees with partial marginalization,
	we propose a variant of the Inside algorithm, the \textsc{Masked Inside} algorithm, 
	that supports different inference operations for different nodes 
	(evaluation for the observed, marginalization for the latent, and rejection for nodes incompatible with the observed) 
	with efficient parallelized implementation, thus significantly speeding up training and inference. 
  Experiments show that our approach achieves the state-of-the-art (SOTA) F1 scores on the ACE2004, ACE2005 dataset, and shows comparable performance to SOTA models on the GENIA dataset. 
  Our approach is implemented at: \url{https://github.com/FranxYao/Partially-Observed-TreeCRFs}.
\end{abstract}

\section{Introduction} 
\label{sec:intro}
Named entity recognition (NER) is a fundamental task in natural language processing \citep{mccallum2003early}. 
Although recent work shows huge success in flat NER with modern neural architectures and pretrained encoders \citep{huang2015bidirectional, devlin2019bert},
NER with \textit{nested structures} is still difficult since simple sequence labeling techniques cannot model these structures \citep{finkel2009nested}. 
Nested NER is also important because entities of nested structures are observed in many domains due to their compositionality
\citep{alex2007recognising} and consequently involved in many real-world applications \citep{kim2003genia}.

\begin{figure}
  \begin{center}		
    \includegraphics[width=3.2in]{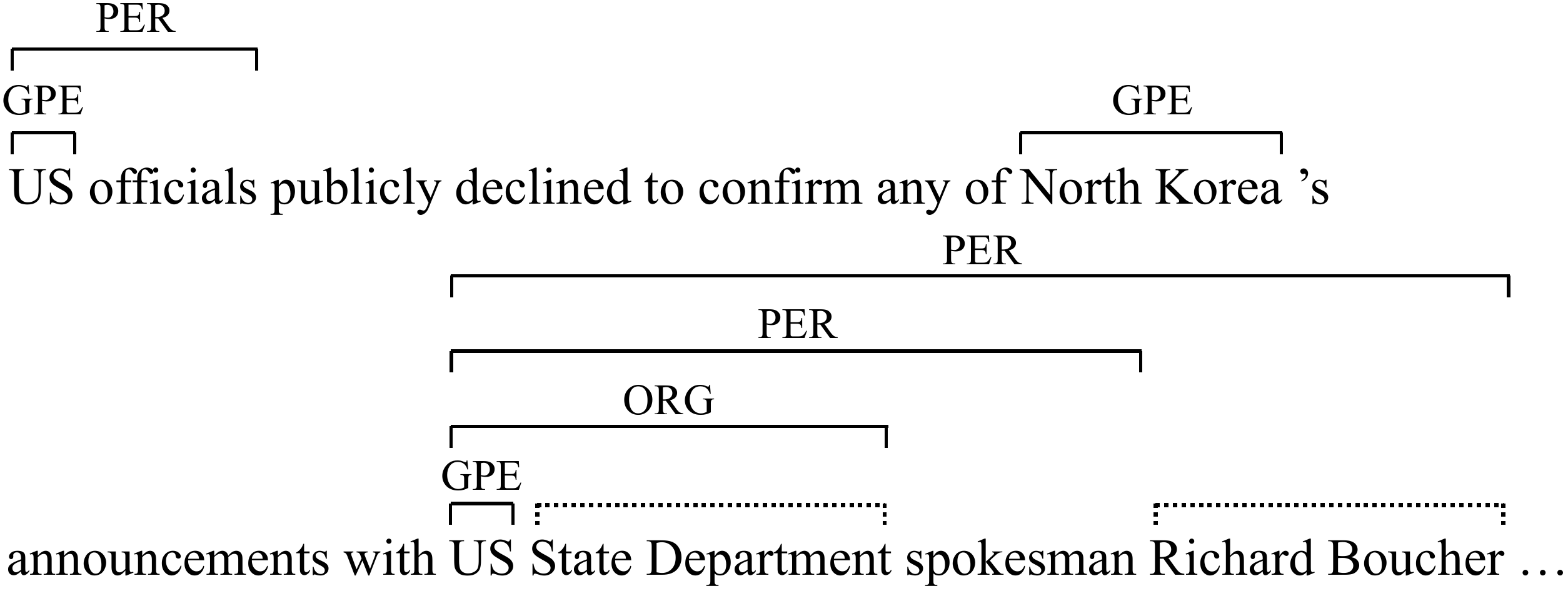}
  \end{center}	
  \caption{\label{fig:ner_example} An example sentence in the ACE dataset with its nested entities. 
  Viewing the nested entity structure as a partially observed tree, 
  a key observation is that other spans without annotation can be modeled as \textit{latent nodes} (dashed lines) in a full tree. 
  This observation motivates us to model nested NER with partially-observed constituency trees.}
\end{figure}

Figure~\ref{fig:ner_example} gives an example sentence with nested entities. 
We observe that an inner entity can be part of an outer entity, which is quite similar to the constituency structure. 
Additionally, the boundaries of nested entities cannot cross. 
These observations motivate us to formulate nested NER as parsing with partially observed constituency trees: 
we can view entities with annotations as \textit{observed constituents}, and assume \textit{a distribution of latent constituents} over spans without annotation. 
For example, \textit{State Department} and \textit{Richard Boucher} can be two possible latent entities in Figure~\ref{fig:ner_example}.

In this work, we propose to model observed and latent entities jointly with a TreeCRF \citep{Zhang2020FastAA, Rush2020TorchStructDS}. 
Specifically, we use a pretrained encoder to obtain word representations, a biaffine scoring mechanism \citep{Dozat2017DeepBA} to obtain log potentials, and a TreeCRF to decode full constituency trees. 
Using TreeCRFs gives the advantage of modeling different types of entities in a probabilistically principled way and properly handling the ambiguities of the latent constituents. 
For optimization, we marginalize all latent constituents out, and maximize the resulting probability of observed partial trees. 

Previously, the application of TreeCRFs for parsing is limited by the cubic time complexity of the Inside algorithm \citep{Eisner2016InsideOutsideAF}.
Recent works show that it is possible to parallelize the Inside algorithm on modern hardware \citep{Rush2020TorchStructDS}. 
While a vanilla Inside algorithm sums over all possible trees, in our setting, we require an Inside-styled partial marginalization which only sums over latent nodes.
To adapt the Inside algorithm to partial summation, we propose a masking method that differentiates different nodes during marginalization. 
Furthermore, to efficiently compute  partial marginalization, we propose a \textsc{Masked Inside} algorithm that performs different inference operations for different types of nodes in a unified masked summation framework. 
We highlight the advantages of the \textsc{Masked Inside} compared with a naive partial marginalization algorithm from two perspectives: 
(a) it is highly batchifiable and parallelizable, allowing us to fully exploit the computational power of modern hardware (like GPUs) and tensor libraries (like Pytorch); 
(b) it is conceptually simple and can be easily implemented by reusing existing implementation of the original Inside algorithm in highly optimized structured prediction libraries (like Torch-Struct \citealt{Rush2020TorchStructDS}). 

We further propose two regularization techniques for TreeCRFs.
Specifically, we propose \textit{potential normalization},  inspired by batch normalization \citep{Ioffe2015BatchNA}, and \textit{structure smoothing}, inspired by label smoothing \citep{Mller2019WhenDL}. 
These two regularizations can be seamlessly integrated with \textsc{Masked Inside}, making their implementation simple and efficient. 

We conduct experiments on three standard benchmark datasets. 
Experimental results show that our approach achieves 86.6, 85.4, and 78.2 scores in terms of F1 on the ACE2004, ACE2005, and GENIA datasets, respectively, which achieves SOTA F1 scores on the ACE2004, ACE2005 dataset, and shows comparable performance to SOTA models on the GENIA dataset. 
We will release the codes for further research. 
Our contributions are:

\begin{itemize}
  \item We propose to formulate nested NER as constituency parsing with partial trees
   and use partially-observed TreeCRFs to jointly model observed and latent nodes. 
  \item We propose the \textsc{Masked Inside} algorithm for efficient partial marginalization and its regularization techniques.
  \item We demonstrate the effectiveness of our proposed methods with extensive experiments. 
\end{itemize}


\section{Related Work}
\label{sec:related_work}
	
\subsection{Nested NER}

It has been a long history of research involving named entity recognition \citep{zhou2002named, mccallum2003early}. 
In the era of deep learning, the LSTM-CRF model achieves very good results in recognizing named entities \citep{huang2015bidirectional,lample2016neural}, especially when equipped with pretrained encoders \citep{Peters2018DeepCW, devlin2019bert}.
\citet{finkel2009nested} point out that named entities are often nested while
traditional sequential labeling models cannot handle the nested structure because they can only assign one label to each token.
Earlier research on nested NER is rule-based \cite{zhang2004enhancing}. 
Recent works in nested NER are in various paradigms as follows:

\noindent \textbf{Hypergraph-based} 
\citet{lu2015joint,katiyar2018nested,wang2018hyper} propose the hypergraph-based method to solve this problem. 
They design a hypergraph to represent all possible nested structures, which guarantees that nested entities can be recovered from the hypergraph tags. 
However, the hypergraph needs to be carefully designed to avoid spurious structures and structural ambiguities which inevitably leads to higher modeling and computational complexity. 
Compared with hypergraph-based approaches, our method tackles ambiguities in a probabilistically principled way by marginalizing all possible latent spans out, and can be implemented easily and efficiently. 

\noindent  \textbf{Transition-based}
Transition-based models are generally similar to shift-reduce parsers with tailored actions for different formalisms. 
\citet{wang2018transition} propose a method to construct nested mentions via a sequence of shift/ reduce/ unary actions. 
\citet{fisher2019merge} propose to form nested structures by merging tokens and/or entities into entities for entity representation. 
Compare with these methods, our approach does not involve the manual labor for designing specialized transition systems, which largely requires domain expertise thus being not generalizable. 
Our partially observed TreeCRFs is more general-purpose and can be flexibly applied to partial trees of different formalisms. 
	

\noindent \textbf{Span-based} 
Another strategy for nested NER is the span-based methods \cite{xu2017local,sohrab2018deep, xia-etal-2019-multi,luan-etal-2019-general, zheng-etal-2019-boundary, tan2020boundary, jue2020pyramid}.
These models first compute the representations for all subsequences in a sentence with tailored neural architectures, 
then classify these spans with locally-normalized scores.
Compared with these models, our TreeCRF can model the dependency for all subsequences with a globally-normalized structured distribution, which consequently leads to clear performance improvements. 

\noindent \textbf{Others} 
There are many other attempts for Nested NER. 
\citet{muis2017labeling} develop a gap-based tagging schema to capture nested structures. 
\citet{lin-etal-2019-sequence} propose a sequence-to-nuggets architecture for nested mention detection. 
\citet{strakova-etal-2019-neural} propose to use a sequence-to-sequence framework to predict the entity label one by one. 
\citet{li-etal-2020-unified} treat NER as a machine reading comprehension task.
Generally, it is hard to study nested NER in a unified framework and we aim to model it in a probabilistically principled way with TreeCRFs. 

	
\subsection{Constituency Parsing with TreeCRFs}
TreeCRFs \citep{Eisner2000BilexicalGA, Eisner2016InsideOutsideAF, Zhang2020FastAA} are well-studied in the parsing literature. 
Before the resurgence of deep learning, their application is limited due to its cubic time complexity. 
Recent work shows that with parallel computation \citep{Zhang2020FastAA,Rush2020TorchStructDS}, they can be efficiently implemented on modern hardware with high-optimized tensor operation libraries, reducing the complexity to at least quadratic time. 

Traditional literature focus on parsing with full annotations.
Although some works study partial annotation, many of them are limited in simulated datasets, e.g., by dropping out certain nodes from fully-annotated trees \citep{Zhang2017DependencyPW, Zhang2020EfficientST}. 
Rather than being simulated, we emphasize that our application of TreeCRFs on Nested NER is a real-world example. 
Moreover, we note that our approach is not limited to nested NER, and an interesting future direction would be applying it for parsing with other types of partial trees. 




\section{Model}
\label{sec:model}	
	

Our model consists of a pretrained encoder, a biaffine scoring module, and a TreeCRF model. 
Given a sentence, we obtain the contextualized representations from the encoder, feed the representations to the biaffine layer to get the log potentials for the TreeCRF model, then use the TreeCRF to decode a constituency tree. 
To calculate the probability of partial trees, we propose the \textsc{Masked Inside} algorithm as a simple, efficient algorithm for partial marginalization. 

\subsection{The Base Biaffine Scoring Architecture}
\label{ssec:base_biaffine}
Given a sentence $x$ as a sequence of words: $x = [x_1, x_2, ..., x_n], x_i \in \mathcal{V}$, $\mathcal{V}$ is the vocabulary and $n$ is the  sentence length. 
We use a base biaffine encoder similar to the biaffine dependency parser in \citet{Dozat2017DeepBA} to predict span scores: 

\begin{align}
  e_1, ..., e_n &= \text{FF}(\text{Enc}(x)) \\ 
  s_{ijk} &= e_i^\intercal U_k^{(1)} e_j + (e_i + e_j)^\intercal U_k^{(2)} + b_k \label{eq:biaffine}
\end{align} 
Where Enc$(\cdot)$ denotes a pretrained encoder, FF$(\cdot)$ denotes a feed-forward network, $e_i$ denotes the contextualized embedding for word $x_i$, $U_k^{(1)}$, $U_k^{(2)}$ and $b_k$ are the parameters for the biaffine scoring mechanism, and $s_{ijk}$ means the score for a constituent spanning from word $x_i$ to $x_j$ (inclusive) with label $k$ where
$i, j \in [1, 2, ..., n], k \in [1, 2, ..., |\mathcal{L}|]$, $\mathcal{L}$ is the set of labels for the constituents. 
We further note $\mathcal{L}$ is a union of observed labels $\mathcal{L}_o$ and latent labels $\mathcal{L}_l$, as we will explain later.

\begin{figure}
  \begin{center}		
    \includegraphics[width=3.25in]{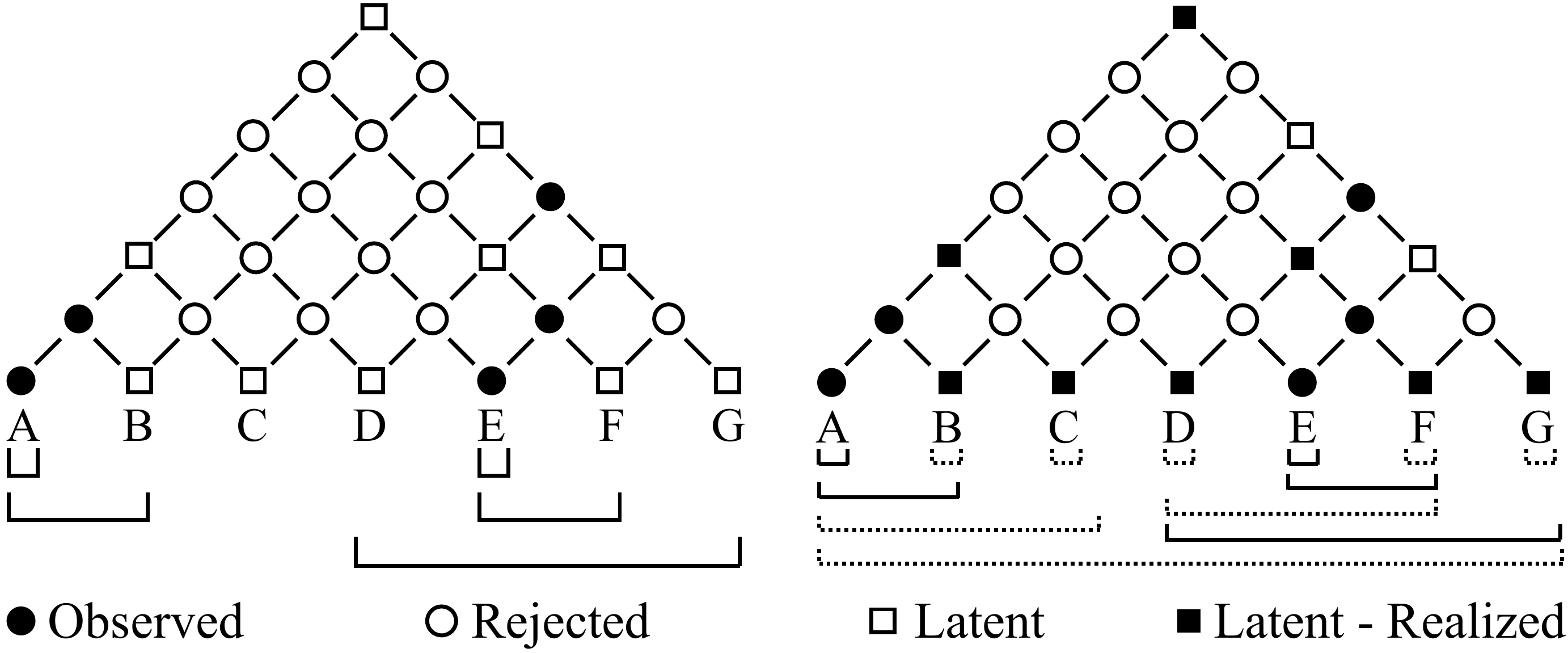}
  \end{center}	
  \caption{\label{fig:example_tree} An example symbol tree.
  Left: observed partial tree (lower) and its corresponding symbol tree (upper) with different types of nodes induced from the observed tree.
  The symbol tree notation further enables us to build different types of masks for the \textsc{Masked Inside} algorithm. 
  Right: a full tree compatible with the left partial tree (lower) by realizing the latent nodes from \scalebox{0.8}{$\square$} to \scalebox{0.8}{$\blacksquare$} (upper). 
  An entity with dashed lines corresponds to a realized latent node \scalebox{0.8}{$\blacksquare$}. 
  A partial tree may correspond to different realized full trees. 
  }
\end{figure}

\subsection{Partially-Observed TreeCRFs}
\label{ssec:po_treecrfs}

A constituency TreeCRF is a probabilistic discriminative model that defines a distribution over constituency trees $T$ given  sentence $x$. 
We represent a labeled constituency tree as a rank-3 binary tensor $T$ where $T_{ijk} = 1$ means that there is a span from word $x_i$ to $x_j$ with label $k$. 
We use the biaffine scores $s_{ijk}$ as the log potentials, and the probability of a tree is given by the Gibbs distribution: 

\begin{align}
  s(T) &= \sum_{ijk} T_{ijk}  s_{ijk} \\
  p(T | x) &= \frac{\exp(s(T))}{Z} \\
  Z &= \sum_T \exp(s(T)) = \textsc{Inside}(s) 
\end{align}
Where $Z$ is the partition function that sums over all possible tree structure $T$, which can be computed exactly with the Inside algorithm \citep{Eisner2016InsideOutsideAF}.

For nested NER, the annotations of trees are incomplete so we only get partial trees. 
With a slight abuse of notation, we still use $T$ to denote partial trees,
and there are locations in $T$ that might be 1 but filled in with 0 because it is not observed (latent). 
To better understand the nature of different nodes in a partial tree, 
we introduce a convenient \textit{symbol tree}  notation $\bar{T}$ given a partial tree $T$. 
$\bar{T}$ is a $n \times n$ matrix with difference types of nodes.
Algorithm \ref{alg:build_mask} gives details for building symbol trees $\bar{T}$ (we delay the discussion about the mask $M$ returned from Algorithm \ref{alg:build_mask} to the next section).
Nodes in $\bar{T}$ can only be one of $\bullet$, \scalebox{0.8}{$\square$} or $\circ$. 
Figure \ref{fig:example_tree} left gives an example of $\bar{T}$. 
A $\bullet$ in $\bar{T}$ means an observed node (a labeled entity) in $T$, 
a $\circ$ means a node that is incompatible with observed nodes because it overlaps with an observed entity 
(e.g., there cannot be a latent entity [BC] because B is already in the observed entity [AB] and the boundaries of entities cannot cross)
and 
a \scalebox{0.8}{$\square$} means a latent node that is possible to be realized in a full tree (e.g., there is a possible entity [ABC] with some latent label). 
We note leaf nodes and the root can only be observed $\bullet$ or latent \scalebox{0.8}{$\square$}. 

\begin{algorithm}[t!]
  \small
  \caption{\label{alg:build_mask} \textsc{Symbol Tree and Mask Construction}}
  \begin{algorithmic}[1]
  
  \State Input: partial tree $T$, $\mathcal{L}_o$ observed labels, $\mathcal{L}_l$ latent labels. 
  \State $\triangleright$ Initialize all nodes as latent \scalebox{0.8}{$\square$}:
  \State \begin{varwidth}[h]{\linewidth} \label{alg:line_latent_mask}
      For all $i, j \in \{1, 2, ..., n\}$: \par
      \hskip\algorithmicindent  $\bar{T}[i, j] = \text{\scalebox{0.8}{$\square$}}$ \par
      \hskip\algorithmicindent  $\forall k_1 \in \mathcal{L}_o, M[i, j, k_1] = 0;\;\; \;\forall k_2 \in \mathcal{L}_l, M[i, j, k_2] = 1$
    \end{varwidth}
  \For{$i, j \gets 1$ to $n$}    
    \If{$\exists k, T_{ijk} = 1$} \Comment{Observe entity $(i, j)$ with label $k$}
      \State $\bar{T}[i, j] = \bullet$ 
      \State $M[i, j, k] = 1$, $\forall m \in \mathcal{L}, m \ne k, M[i, j, m] = 0$ \label{alg:line_observed_mask}
      \State $\triangleright$ For all spans with crossed boundaries with $(i, j)$:
      \State \begin{varwidth}[h]{\linewidth} \label{alg:line_rejected_mask_1}
        $\forall (i', j'), i' < i \; \&\; i \le j' < j$: \par
          \hskip\algorithmicindent  $\bar{T}[i', j'] = \circ, \;\;\forall m \in \mathcal{L}, M[i', j', m] = 0$ \par
        \end{varwidth}
      \State \begin{varwidth}[h]{\linewidth} \label{alg:line_rejected_mask_2}
        $\forall (i', j'),  i < i' \le j \; \&\; j < j'$: \par
          \hskip\algorithmicindent  $\bar{T}[i', j'] = \circ, \;\;\forall m \in \mathcal{L}, M[i', j', m] = 0$ \par
        \end{varwidth}
    \EndIf
  \EndFor

  \State Return: symbol tree $\bar{T}$, mask $M$
  
\end{algorithmic}
\end{algorithm} 

Given the partial tree $T$, a full tree $\tilde{T}$ compatible with $T$ can be constructed by realizing \scalebox{0.8}{$\square$} to \scalebox{0.8}{$\blacksquare$} in $\bar{T}$
(Figure \ref{fig:example_tree} right. A \scalebox{0.8}{$\blacksquare$} in the upper part denotes a realized latent span corresponding to an entity with a dashed line in the lower part). 
We further denote the set of labels for latent spans $\mathcal{L}_{l}$ as opposed to the labels for observed entities $\mathcal{L}_{o}$. 
We restrict the labels for the latent spans to be within $\mathcal{L}_{l}$. 
For example, the labels for the constituents [ABC] and [DEF] are only allowed to be in $\mathcal{L}_{l}$ because they are latent. 
This separation would allow us to decode partial trees during inference by dropping out entities with latent labels. 
Since there are multiple ways to complete a partial tree, we use $\tilde{\mathcal{T}}$ to denote the set of full trees completed from $T$. 

To train the TreeCRF with partial trees, 
we maximize the conditional probability of $p(T| x)$ computed by marginalizing all latent nodes \scalebox{0.8}{$\square$} out:

\begin{align}
  &\log p(T | x) = s(T) - \log Z  \label{eq:log_prob} \\ 
  &s(T) = \log \sum_{\tilde{T} \in \tilde{\mathcal{T}}} \exp( s(\tilde{T})) \label{eq:summation}
\end{align}
This objective could equivalently be viewed as the average probability of the observed partial tree $T$ over its all possible compatible full trees in $\tilde{\mathcal{T}}$.

  

\begin{algorithm}[t]
  \small
  \caption{\label{alg:partial_summation} \textsc{Inside for Partial Marginalization}}
  \begin{algorithmic}[1]
  
  \State Input: Scores $s$, partial tree $T$ and its corresponding $\bar{T}$
  \For{$i \gets 1$ to $n$}  
    \If{$\bar{T}[i, i] = \bullet$} \Comment{Observed leaf}
      \State $\exists k \in \mathcal{L}_o, T_{iik} = 1, \beta[i, i, k] = \exp(s_{iik})$   \label{alg:line_observed_1}  
      \State $\forall m \ne k, \beta[i, i, m] = 0 $
    \ElsIf{$\bar{T}[i, i] = \text{\scalebox{0.8}{$\square$}}$} \label{alg:line_latent_1} \Comment{Latent leaf} 
      \State $\forall k \in \mathcal{L}_o, \beta[i, i, k] = 0 $ \label{alg:line_latent_o_1}
      \State $\forall k \in \mathcal{L}_l, \beta[i, i, k] = \exp(s_{iik}) $ \label{alg:line_latent_l_1}
    \EndIf
  \EndFor

  \For{$d \gets 1$ to $n - 1$}       
    \For{$i \gets 1$ to $n - d$}       
      \State $j = i + d$
      \If{$\bar{T}[i, j] = \bullet$}  \Comment{Observed}
        \State $\exists k \in \mathcal{L}_o, T_{ijk} = 1$ 
        \State \begin{varwidth}[h]{\linewidth}
          $\beta[i, j, k] = \exp(s_{ijk}) \cdot$ \par
            \hskip\algorithmicindent $ \sum_{l = i}^{j - 1} \sum_{k_1, k_2 \in \mathcal{L}} \beta[i, l, k_1] \beta[l + 1, j, k_2]$
          \end{varwidth} \label{alg:line_observed_2}
        \State $\forall m \ne k, \beta[i, j, m] = 0 $
      \ElsIf{$\bar{T}[i, j] = \text{\scalebox{0.8}{$\square$}}$} \label{alg:line_latent_2} \label{alg:line_latent_2} \Comment{Latent}
        \State \begin{varwidth}[h]{\linewidth} \label{alg:line_latent_l_2}
          $\forall k \in \mathcal{L}_l, \beta[i, j, k] = \exp(s_{ijk}) \cdot$ \par
            \hskip\algorithmicindent $ \sum_{l = i}^{j - 1} \sum_{k_1, k_2 \in \mathcal{L}} \beta[i, l, k_1] \beta[l + 1, j, k_2]$
          \end{varwidth}
        \State $\forall k \in \mathcal{L}_o, \beta[i, j, k] = 0 $ \label{alg:line_latent_o_2}
      \ElsIf{$\bar{T}[i, j] = \circ$} \label{alg:line_rejected} \Comment{Rejected}
        \State $\forall k \in \mathcal{L}, \beta[i, j, k] = 0 $ \label{alg:line_sum_end}
      \EndIf
    \EndFor
  \EndFor

  \If{$\bar{T}[1, n] = \bullet$}  \label{alg:line_observed_3}  \Comment{Observed root}
    \State $\exists k \in \mathcal{L}_o, T_{1nk} = 1$. Return $s(T) = \beta[1, n, k]$
  \ElsIf{$\bar{T}[1, n] = \text{\scalebox{0.8}{$\square$}}$} \label{alg:line_latent_3} \Comment{Latent root}
    \State Return $s(T) = \log(\sum_{k \in \mathcal{L}_l}\beta[1, n, k])$ \label{alg:line_latent_l_3}
  \EndIf
  
\end{algorithmic}
\end{algorithm}

\subsection{Masked Inside for Efficient Partial Marginalization}
\label{ssec:masked_inside}
To compute the partial marginalization in equation~(\ref{eq:summation}), we give a tailored Inside algorithm that supports different inference operations for different nodes.
As is shown in Algorithm \ref{alg:partial_summation}, during the summation process,
if the current node is:
(a) an observed~$\bullet$, then we evaluate (add) its corresponding score (line \ref{alg:line_observed_1} and \ref{alg:line_observed_2}); 
(b) a latent~\scalebox{0.8}{$\square$} whose label can only be in $\mathcal{L}_l$.
So we reject (do not add) all the scores corresponding to observed labels $\mathcal{L}_o$  (line \ref{alg:line_latent_o_1} and \ref{alg:line_latent_o_2}), and sum over all scores corresponding to latent labels $\mathcal{L}_l$ for this node (line \ref{alg:line_latent_l_1} and \ref{alg:line_latent_l_2}); 
(c) a rejected~$\circ$, we reject all scores corresponding to this node (line \ref{alg:line_sum_end}). 

However, a naive implementation of Algorithm~\ref{alg:partial_summation} can be inefficient with $O(n^3)$ complexity. 
Such inefficiency has previously restricted the application of TreeCRFs in parsing literature. 
More severely, Algorithm~\ref{alg:partial_summation} does not support batch computation over sentences, making it more impractical.
Recent works show that, for the original Inside algorithm, it is possible to parallelize it on modern hardware architectures with efficient batch computation, reducing the complexity from $O(n^3)$ to at least $O(n^2)$, and could further be $O(n\log n)$\footnote{
The complexity discussed here does not include the summation over $k_1, k_2$ in line~\ref{alg:line_masked_sum}. If the summation over $l$ is implemented with a binary tree shaped summation then the complexity is $O(n \log n)$, if it is sequential then the complexity is $O(n^2)$. The optimization of this summation is usually implemented inside tensor computation libraries.
}
\citep{Rush2020TorchStructDS, Zhang2020FastAA}. 
It would be ideal if we could use similar batchification techniques for Algorithm~\ref{alg:partial_summation}, which motivates our proposed \textsc{Masked Inside} algorithm for efficient partial marginalization. 

  

\begin{algorithm}[t!]
  \small
  \caption{\label{alg:masked_inside} \textsc{Masked Inside}}
  \begin{algorithmic}[1]
  
  \State Input: Scores $s$, mask $M$
  \For{$i \gets 1$ to $n$}    
    \State $\beta[i, i, k] = M[i, i, k] \cdot \exp(s_{iik})$ \label{alg:line_masked_leaf} \Comment{Masked leaf scores} 
  \EndFor
  \For{$d \gets 1$ to $n - 1$}       
    \State \begin{varwidth}[h]{\linewidth} 
      Parallelization on $i$, tensor operation on $l, k, k_1, k_2$ \label{alg:line_parallel} \par
        \hskip\algorithmicindent  $1 \le i \le n - d$; $\;\; j = i + d$;  $\;\; k, k_1, k_2 \in \{1, ..., |\mathcal{L}|\}$
      \end{varwidth}
    \State \begin{varwidth}[h]{\linewidth} \label{alg:line_masked_sum}
      $\beta[i, j, k] = (M[i, j, k] \exp(s_{i j k})) \; \cdot$ \Comment{Masked scores} \par
        \hskip\algorithmicindent $\sum_{l = i}^{j - 1} \sum_{k_1, k_2 \in \mathcal{L}} \beta[i, l, k_1] \beta[l + 1, j, k_2]$
      \end{varwidth}
  \EndFor

  \State Return: $s(T) = \log(\sum_{k \in \mathcal{L}} \beta[1, n, k])$
  
\end{algorithmic}
\end{algorithm}

As is shown in Algorithm~\ref{alg:masked_inside}, the key insight of \textsc{Masked Inside} is that \textit{all if-else statements for partial summation in Algorithm~\ref{alg:partial_summation} 
can be re-written in a unified masked summation} (line~\ref{alg:line_masked_leaf} and \ref{alg:line_masked_sum} in Algorithm~\ref{alg:masked_inside}) with a pre-computed mask $M$ from Algorithm~\ref{alg:build_mask}. 
To be specific, in Algorithm~\ref{alg:build_mask}: 
(a) for an observed node~$\bullet$, we mask out all scores except the score of its observed label (line~\ref{alg:line_observed_mask}), 
which corresponds to lines~\ref{alg:line_observed_1}, \ref{alg:line_observed_2} and \ref{alg:line_observed_3} in Algorithm~\ref{alg:partial_summation}; 
(b) for a rejected node~$\circ$, we mask out all its scores (lines \ref{alg:line_rejected_mask_1}-\ref{alg:line_rejected_mask_2}), which corresponds to line~\ref{alg:line_rejected} in Algorithm~\ref{alg:partial_summation}; 
(c) for a latent node~\scalebox{0.8}{$\square$}, we mask out the scores for all observed labels $\mathcal{L}_o$, and retain scores for all latent labels $\mathcal{L}_l$ (line~\ref{alg:line_latent_mask}), which corresponds to lines~\ref{alg:line_latent_1}, \ref{alg:line_latent_2}, and \ref{alg:line_latent_3} in Algorithm~\ref{alg:partial_summation}. 
Applying different masks to Algorithm~\ref{alg:masked_inside}, we can recover all if-else statements in Algorithm~\ref{alg:partial_summation}. 
As two special cases, if all masks are 1 (i.e., not masked), we recover the original Inside algorithm; if the masks are constructed from a full tree, we recover the original bottom-up evaluation. 

\begin{table*}[t]
  \centering
  \begin{tabular}{@{} l  ccc  ccc  ccc @{}}
    \toprule
    & \multicolumn{3}{c}{ACE2004} & \multicolumn{3}{c}{ACE2005} & \multicolumn{3}{c}{ GENIA} \\ 
    &Train & Dev & Test &Train & Dev & Test &Train & Dev & Test \\ 
    \hline			
    \# sentences & 7,078 & 859 & 922 & 7,194 & 969 &1,047 & 14,836 & 1,855 & 1,855 \\ 			
    {with nested entities } & 2,691 & 290 & 377 & 2,691 & 338 & 330 & 3,199 & 362 & 448 \\			
    \# entities & 22,172 &2,510 & 3,024 & 24,441 & 3,200 & 2,993 & 46,473 & 5,014 & 5,600 \\			
    {\# nested entities} & 10,080 &1,086 & 1,410 &9,389 & 1,112& 1,118 & 8,337 & 903 & 1,217 \\			
    avg length & 20.38 & 20.69 & 20.96 & 19.21 & 18.93 &17.19 &30.13 & 29.17 &30.48 \\
    \bottomrule
  \end{tabular}
  \caption{Statistics of ACE2004, ACE2005, and GENIA datasets.}
  \label{dataset}
\end{table*}

	\begin{table*}[t]
		\begin{center}
			\begin{tabular}{@{}l c c c c c c c c c @{}}
				\toprule
				& \multicolumn{3}{c}{ACE2004} 	& \multicolumn{3}{c	}{ACE2005} & \multicolumn{3}{c}{GENIA} \\ 
				Model & P & R & F1 & P & R & F1 & P & R & F1 \\
				\hline						
				LSTM-CRF~\cite{lample2016neural} & 71.3 & 50.5 & 58.3 & 70.3 & 55.7 & 62.2 & 75.2 & 64.6 & 69.5 \\
				FOFE(c=6)~(Xu et al. 2017) & 68.2& 54.3 & 60.5 & 76.5 & 66.3 & 71.0 & 75.4 & 67.8 & 71.4 \\
				Transition~\cite{wang2018transition} & 74.9 &71.8 & 73.3& 74.5 & 71.5 & 73.0 & 78.0 & 70.2 & 73.9 \\
				Cascaded-CRF~(Ju et al. 2018) &-&-& - & 74.2 & 70.3 & 72.2 & 78.5 & 71.3 & 74.7 \\		
				SH(c=n)~\cite{wang2018hyper} & 77.7 &72.1&74.5 & 76.8 & 72.3 & 74.5 & 77.0 & 73.3 & 75.1 \\  				
				ML~\cite{fisher2019merge} &-&-&- & 75.1& 74.1 &74.6 & - &- &- \\
				BENSC~\cite{tan2020boundary} & 78.1 & 72.8 & 75.3 & 77.1 & 74.2 & 75.6 & 78.9 & 72.7 & 75.7\\
				Pyramid \cite{jue2020pyramid} & 81.1 & 79.4 & 80.3 & 80.0 & 78.9 & 79.4 & 78.6 & 77.0 & 77.8 \\
				\hline
				\multicolumn{8}{@{}l}{with Pretrained LM } \\	
				MGNER (ELMo)~\cite{xia-etal-2019-multi}& 81.7& 77.4& 79.5 & 79.0 & 77.3 & 78.2 &-&-&-  \\
				ML (ELMo) \cite{fisher2019merge} &- &- &-& 79.7 & 78.0 & 78.9 &- &- &- \\
				ML (BERT) \cite{fisher2019merge} &- &- &- & 82.7 & 82.1 & 82.4 &- &- &- \\
				Seq2seq \cite{strakova-etal-2019-neural} &- &- & 84.3 &- &- & 83.4 &- &- & 78.2 \\
				BENSC (BERT) \cite{tan2020boundary} & 85.8 & 84.8 & 85.3 & 83.8& 83.9 & 83.9 & 79.2 & 77.4 & 78.3 \\
				Pyramid (BERT) \cite{jue2020pyramid} & 86.1 & 86.5 & 86.3 & 84.0 & 85.4 & 84.7 & 79.5 & 78.9 & 79.2 \\
				\hline
				\multicolumn{8}{@{}l}{with Additional Supervision } \\
				DYGIE \cite{luan-etal-2019-general} &- &- & 84.7 &- & - & 82.9 & - & - & 76.2 \\ 
				\citet{yu-etal-2020-named}  & 87.3 & 86.0 & 86.7 & 85.2 & 85.6 & 85.4 & 81.8 & 79.3 & 80.5 \\
				BERT-MRC \cite{li-etal-2020-unified} & 85.0 &  86.3 & 86.0  & 87.2 & 86.6 & 86.9 & 85.2 & 81.1 & 83.8 \\ 
				\hline
				\multirow{2}{5cm}{\bf PO-TreeCRFs (ours)} & 86.7 & 86.5 & 86.6 &  84.5 	  &  86.4 & 85.4 & 78.2 & 78.2 & 78.2 \\
				& \small{$\pm$0.4}  & \small{$\pm$0.4}  & \small{$\pm$0.3}  & \small{$\pm$0.4}  & \small{$\pm$0.2}  & \small{$\pm$0.1}  & \small{$\pm$0.7}  & \small{$\pm$0.8}  & \small{$\pm$0.1}   \\
				\hline
				\multicolumn{8}{@{}l}{\bf PO-TreeCRFs Ablation Study} \\	
				Change Biaffine to Bilinear & 86.0 & 86.7 & 86.4 & 83.0 & 86.5 & 84.7 & 79.9 & 75.5 & 77.6   \\
				W/o. Structure Smoothing & 86.1 & 86.4 & 86.2 & 83.5 & 85.8 & 84.6 & 78.7 & 76.5 & 77.6 \\
				W/o. Potential Normalization and Structure Smoothing & 86.0 & 85.3 & 85.7 & 82.7 & 86.2 & 84.4 & 76.5 & 78.1 & 77.3 \\
				W/o. TreeCRFs & 84.4 & 85.4 & 84.9 & 82.0 & 86.4 & 84.1 & 80.5 & 74.5 & 77.4 \\ 
				\bottomrule
			\end{tabular}		
		\end{center}
		\caption{Main results and ablation studies on three datasets. We report the average scores of 5 runs for main results. } 
		\label{result}
	\end{table*}

As an efficient alternative for Algorithm~\ref{alg:partial_summation}, the advantages of Algorithm~\ref{alg:masked_inside} is that it is (a) conceptually much simpler and (b) highly parallelizable.
The later allows us to fully exploit the computational power of modern hardware architectures (like GPUs) and highly optimized tensor operation libraries (like Pytorch).
We note that it the parallelization on $i$ in Algorithm~\ref{alg:masked_inside} that reduces the complexity to at least $O(n^2)$, 
Furthermore, as an equivalent implementation to Algorithm~\ref{alg:masked_inside}, we can apply masks to scores in the logarithm scale\footnote{In our implementation, we use $-10^6$ to substitute the undefined $\log 0$ for numerical stability.}
then feed the masked scores to a normal Inside algorithm (rather than multiplying the masks inside Algorithm~\ref{alg:masked_inside}): 

\begin{align}
  s(T) &= \textsc{MaskedInside}(s, M) \nonumber \\
  &= \textsc{Inside}( \log M + s) \label{eq:masked_inside}\\ 
  p(T | x) &= \frac{\exp(s(T))}{Z} \label{eq:log_prob_efficient}
\end{align}
In practice, we compute the masks $M$ in the data-processing stage.
For training, by reusing existing implementations of the Inside algorithm in highly-optimized structured prediction libraries like Torch-Struct~\citep{Rush2020TorchStructDS},
we can \textit{implement the \textsc{Masked Inside} with one single line of code}, as is shown in equation~(\ref{eq:masked_inside}), thus significantly reducing the implementation complexity required by Algorithm~\ref{alg:partial_summation}.

\subsection{Regularization} 
\label{ssec:regularization}
We propose two regularization techniques for TreeCRFs: (a)
\textit{potential normalization}, which is inspired by batch normalization \citep{Ioffe2015BatchNA}, and (b) \textit{structure smoothing}, which is inspired by label smoothing  \citep{Mller2019WhenDL}. 
Potential normalization (PN) is simple: we normalize the scores $s$ to an empirical distribution of zero mean and one variance. 
The difference with batch-norm (BN) is that we apply PN at an instance-level, rather than a batch-level. 
In our experiments, we observe that PN gives a slightly better convergence. 

Structure smoothing regularizes TreeCRFs by putting a small portion of weights to nodes that are marginalized out. 
Specifically, during the partial marginalization, instead of using a zero mask that does not include the weights of rejected nodes, we change the mask to a small value $\epsilon$

\begin{align}
  M[i, j, k] = 0 \to M[i, j, k] = \epsilon \quad \text{for rejected $\circ$}
\end{align}
This would effectively add the weights of all rejected nodes to $s(T)$ with a multiplier $\epsilon$. 
This is similar to label smoothing which adds a small portion of weights to all labels other than the target label. 
The reason that we call it structure smoothing is that it not only smooths over the labels, but also smooths over different tree structures\footnote{Additionally, if we randomize $\epsilon$ and make them i.i.d. Gumbel samples, we recover the Stochastic Softmax Trick~\citep{paulus2020gradient}, which generalize Gumbel-Softmax to combinatorial structures.}. 
We further observe that structure smoothing should be based on potential normalization for numerical stability, otherwise it does not converge in our experiments. 
The implementation of structure smoothing is still easy and aligns with previous discussions about equation~(\ref{eq:masked_inside})
as one only needs to change the zeros in $M$ to $\epsilon$.

\subsection{Training and Inference} 
\label{ssec:training_inference}
During training,
we maximize the log conditional probability $\log p(T|x)$ efficiently computed by equation~(\ref{eq:masked_inside}) and~(\ref{eq:log_prob_efficient}).
During inference, we use CKY decoding to decode a full tree with the maximum probability. 
We only include nodes whose labels are in the observed label set $\mathcal{L}_o$, and dismiss nodes whose labels are in the latent set $\mathcal{L}_l$.
This would allow us to decode nested entities (partial trees) for evaluation. 



\section{Experiments}
\label{sec:exp}
We conduct experiments on three standard benchmark datasets. 
We show that our proposed approach achieves SOTA performance.
We further conduct detailed error analysis, case study, and time complexity analysis.

\subsection{Datasets}
We conduct experiments on the ACE2004, ACE2005 \cite{doddington2004automatic}, and GENIA \cite{kim2003genia} datasets.
There are seven types of entities as `FAC', `LOC', `ORG', `PER', `WEA', `GPE', `VEH' in the ACE datasets and five types of entities as `G\#DNA', `G\#RNA', `G\#protein', `G\#cell\_line', `G\#cell\_type' in the GENIA dataset. 
The statistics of these datasets are shown in Table~\ref{dataset}.

\subsection{Implementation Details}

We use variants of BERT \cite{devlin2019bert} to encode sentences. For the ACE2004 and ACE2005 datasets, we use the \textit{bert-large-cased} checkpoint. For GENIA, we use \textit{BioBERT v1.1} \cite{lee2020biobert}. 
As words in the sentence are divided into word pieces, 
we use the representation of the first piece to represent each word after BERT encoding. 
The parameter in BERT is also trainable. 
We use AdamW optimizer with the learning rate~{2e-5} on ACE2004 dataset and~\text{3e-5} on ACE2005 and GENIA dataset. 
The $\epsilon$ used for structure smoothing is~{0.01} on ACE2004 dataset and~{0.02} on ACE2005 and GENIA dataset. 
We apply~{0.2} dropout after BERT encoding.
Denote the hidden size of the encoder as $h$ ($h = 1024$ for BERT Large, and $768$ for BioBERT). 
We apply two feed-forward layers before the biaffine scoring mechanism, with $h$ and $h/2$ hidden size, respectively.
Consequently, the size of biaffine matrix is $ h/2 \times h/2$. 
We set the size of latent labels $\mathcal{L}_l$ to 1 as in our preliminary experiments we find out the performance does not differ significantly with more latent labels. 

\subsection{Baselines}

Below we list our baseline models with comparable  settings. 
We also include the results of models that use additional supervision, which are not directly comparable to ours. 

\begin{description}
	\item [LSTM-CRF] is a classical baseline for NER. This model cannot solve the problem of nested entities \cite{lample2016neural}. 
	\item [FOFE] is a span-based method that classifies over all sub-sequences of a sentence with a fixed-size forgetting encoding \cite{xu2017local}.	
	\item [Transition] is a shift-reduce based system that learns to construct the nested structure in a bottom-up manner through an action sequence \cite{wang2018transition}.
	\item [Cascaded-CRF] applies several stacked CRF layers to recognize nested entities at different levels in an inside-out manner \cite{ju2018neural}.
	\item [SH] improves LH \cite{katiyar2018nested} by considering the transition between labels to alleviate labeling ambiguity of hypergraphs \cite{wang2018hyper}.
	\item [MGNER] first applies the Detector to generate possible spans as candidates and then applies a Classifier for the entity type \cite{xia-etal-2019-multi}.
	\item [Merge and Label (ML)] first merges tokens and/or entities into entities forming nested structures and then labels entities to corresponding types \cite{fisher2019merge}.
	\item [Seq2seq] is under a encoder-decoder framework to predict the entity one by one \cite{strakova-etal-2019-neural}.
	\item [BENSC] is a span-based method that incorporates a boundary detection task for multitask learning~\cite{tan2020boundary}.
	\item [Pyramid] is the state-of-the-art method without external supervision. It recursively inputs tokens and regions into flat NER layers for span representations \cite{jue2020pyramid}.
\end{description}

\begin{figure*}[t]
  \begin{center}		
    \includegraphics[width=0.9\textwidth]{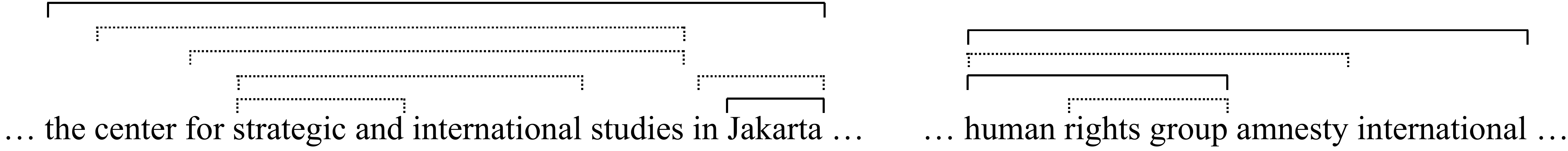}
  \end{center}	
  \caption{\label{fig:inferred_trees} 
  Inferred latent tree structure examples. Solid lines: predicted entities; dashed lines: realized latent constituents. 
  Although there are spans that are not that meaningful like \textit{rights group}, we still observe meaningful inferred spans like \textit{for strategic and international studies} and \textit{in Jakarta}.
  }
\end{figure*}

\subsection{Results}
Table~\ref{result} shows the overall results on ACE2004, ACE2005, and GENIA. 
We primarily compare our model with the Pyramid(BERT) model \cite{jue2020pyramid}, as it achieves SOTA scores without additional supervision signals. 
As we believe there are still rooms for further performance improvements, e.g., to use a more powerful, larger encoder (like GPT3 \citealp{Brown2020LanguageMA}) or to use more ensemble methods, e.g., to ensemble FLAIR \citep{Akbik2018ContextualSE} and other pretrained encoders, to standardize the comparison and validate the effectiveness of TreeCRFs, we restrict the encoder to be BERT. 
We denote our partially-observed TreeCRF as PO-TreeCRF. 
Our PO-TreeCRF achieves 86.6, 85.4, and 78.2 scores in terms of $F_1$ on the ACE2004, ACE2005, and GENIA datasets, respectively, which achieves the state of the art F1 scores on the ACE2004, ACE2005 dataset, and shows comparable performance to Pyramid(BERT) on the GENIA dataset. 

We further emphasize the evaluation of nested NER is not strictly standardized and there are more or less differences across different works. 
As is in Table~\ref{result}, there are models that show improvements with additional information. 
Specifically, DYGIE \cite{luan-etal-2019-general} uses the OntoNotes annotations for better coreference resolution. 
\citet{yu-etal-2020-named} train and evaluate their model at the paragraph level which gives a better coreference resolution performance.
Their work is under a different setting as the training and evaluation of most works are at the sentence level. 
BERT-MRC \cite{li-etal-2020-unified} takes annotation guideline notes as references to construct queries, which is strong supervision and the corresponding corpus are not always easy to obtain. 
As we try to align our evaluation with the majority of literature, the models mentioned above are not directly comparable to our method.



\begin{table}[t!]
  \small
  \begin{center}
    \begin{tabular}{@{} l ccccccc @{}}
      \toprule
      & PER & LOC & ORG & GPE & FAC & VEH & WEA \\
      \hline
  $\rho$ & 0.58 & 0.02 & 0.17 & 0.14 & 0.05 & 0.03 & 0.02 \\
      \hline
 PER & 0.93 & 0.00 & 0.07 & 0.05 & 0.01 & 0.03 & 0.00 \\ 
LOC & 0.00 & 0.72 & 0.00 & 0.02 & 0.05 & 0.00 & 0.00 \\
ORG & 0.02 & 0.02 & 0.77 & 0.00 & 0.02 & 0.01 & 0.00 \\
GPE & 0.00 & 0.06 & 0.03 & 0.84 & 0.03 & 0.00 & 0.00 \\
FAC & 0.00 & 0.02 & 0.01 & 0.00 & 0.70 & 0.03 & 0.00 \\
VEH & 0.00 & 0.00 & 0.00 & 0.00 & 0.01 & 0.67 & 0.00 \\
WEA & 0.00 & 0.00 & 0.00 & 0.00 & 0.00 & 0.01 & 0.84 \\
Latent & 0.02 & 0.15 & 0.08 & 0.05 & 0.12 & 0.15 & 0.10 \\
None & 0.03 & 0.04 & 0.04 & 0.03 & 0.06 & 0.10 & 0.06 \\

      \bottomrule
    \end{tabular}		
  \end{center}
  \caption{\label{tab:error} Error distribution on ACE2005. 
  Rows: entities predicted by our model. None denotes the entities that are not predicted. Columns: labeled entities.
  Numbers are normalized by columns.
  $\rho$ denotes the prior distribution of labels. 
  } 
\end{table}


\subsection{Ablation Study}
Table~\ref{result} lowest rows show the results of ablation study. 
We note that structure smoothing should be based on potential normalization otherwise the model does not converge. 
By without TreeCRFs we mean to use the biaffine scorer and normalize the scores locally, similar to \citet{yu-etal-2020-named}. 
Not using TreeCRFs would lead to the largest performance drop which demonstrates the effectiveness of global normalization with TreeCRFs. 
We also observe performance drop when we do not use potential normalization and structure smoothing which validates their effectiveness for regularizing TreeCRFs.

\subsection{Error Analysis}
In our experiments, we find out the recall for unlabeled spans is 96.3 on ACE2005, 
which means that the spans for most entities are correctly covered, 
and it is their labels that are more difficult to predict. 
To see which labels are more prone to errors, we report the error distribution in Table~\ref{tab:error}. 
We see that the VEH, FAC, and LOC are the top three classes prone to errors as they are extremely imbalanced  (0.03, 0.05, and 0.02 respectively), and many of them are predicted as latent.
This indicates that a future direction is to adapt the TreeCRF to imbalanced labels. 
We leave it to future work.

\subsection{Case Study}
Figure~\ref{fig:inferred_trees} gives examples of inferred latent tree structures compatible with predicted entities.  
As the learning of latent structures is completely unsupervised, we may not expect that the inferred subtrees should align with human intuition, and we do observe some spans that are not that interpretable like \textit{rights group}. 
However, we still observe some meaningful constituents like \textit{for strategic and international studies} and \textit{in Jakarta}, 
which indicates that our approach is indeed learning meaningful tree structures to a certain extent.
We note there are also related unsupervised grammar induction works with TreeCRFs \citep{Kim2019UnsupervisedRN, Kim2019CompoundPC},
and we leave the application of our model to grammar induction to future work. 

\begin{table}[t!]
  \begin{center}
    \small
    \begin{tabular}{@{} l c c c @{}}
      \toprule 
      Method & Inside (Vanilla) & \textsc{Masked Inside} & Biaffine \\
      \hline
      GPU Time & 14m58s & 3m20s & 2m27s \\
      CPU Time & 2h5m & 24m & 22m10s \\
      Complexity & $O(n^3)$ & $O(n \log n)$ & $O(1)$ \\ 
      \bottomrule
    \end{tabular}		
  \end{center}
  \caption{\label{tab:inf_speed} Time for training one epoch on ACE2004. 
  GPU Nvidia P100, CPU Intel 2.6Hz quad-core i7.} 
\end{table}

\subsection{Time Complexity}
Table~\ref{tab:inf_speed} shows the speed for training different models. 
We primarily focus on GPU time, but also report CPU time. 
The base Biaffine model is similar to the model in \citet{yu-etal-2020-named} which uses locally normalized scores, instead of using a TreeCRF.
This model eliminates the complexity of the Inside algorithm and can be computed in $O(1)$ time. 
which can be viewed as an upper bound of time complexity. 
Thanks to the masking mechanism that is compatible with parallelization and tensor operations, our  \textsc{Masked Inside} is significantly quicker than a vanilla implementation of Inside for partial marginalization, and is close to the base Biaffine in practice.

\section{Conclusion}
\label{sec:conclusion}
In this work, we propose to view nested entity structures as partially observed constituency trees, and model it with partially observed TreeCRFs. 
We use a pretrained encoder and a biaffine scoring module to predict the log potentials, then use the TreeCRF to decode the entities. 
We give a detailed discussion of different nodes within partial trees and their corresponding inference operations during partial marginalization. 
To facilitate efficient computation with modern hardware and tensor libraries, we propose the \textsc{Masked Inside} algorithm that is conceptually simple and practically efficient. 
We demonstrate the effectiveness and efficiency of our approach with extensive experiments. 

\section{ Acknowledgments}

We greatly thank all anonymous reviewers for their helpful comments. We also thank Yijia Liu, Yu Zhang, Yue Zhang, Rui Wang, and Ningyu Zhang for helpful discussions.

\bibliography{aaai21}

\appendix
\section{Appendix}
\begin{table*}[!htbp]
	\begin{center}
		\begin{tabular}{@{}l c c c c c c c c c @{}}
			\toprule
			& \multicolumn{3}{c}{ACE2004} 	& \multicolumn{3}{c	}{ACE2005} & \multicolumn{3}{c}{GENIA} \\ 
			 & P & R & F1 & P & R & F1 & P & R & F1 \\
			\hline						
			Run 1 & 86.94 & 86.20 & 86.57 & 84.82 & 86.33 & 85.57 & 77.31 & 79.31 & 78.29 \\
		    Run 2 & 86.92 & 86.63 & 86.77  & 84.80 & 86.42 & 85.61 & 78.92 & 77.58 & 78.24 \\ 
		    Run 3 & 86.83 & 85.97 & 86.40 & 83.92 & 86.76 & 85.31 & 77.96 & 78.39 & 78.17 \\ 
		    Run 4 & 86.14 & 86.63 & 86.38  & 84.16 & 86.52 & 85.33 & 77.85 & 78.62 & 78.24 \\ 
		    Run 5 & 87.03 & 87.09 & 87.06 & 84.62 & 86.09 & 85.35 & 79.00 & 77.29 & 78.14 \\ 
			\bottomrule
		\end{tabular}		
	\end{center}
	\caption{\label{tab:full_result} Detailed Results of 5 Runs. The numbers does not vary much, which means that our model performance is not sensitive to random initialization.}
\end{table*}

\begin{table*}
  \begin{center}
    \begin{tabular}{@{} l c c c c c c c c c @{}}
      \toprule
        & \multicolumn{3}{c }{ACE2004} 	& \multicolumn{3}{c }{ACE2005} & \multicolumn{3}{c }{GENIA}  \\ 
       \# Latent Labels & P & R & F1 & P & R & F1 & P & R & F1 \\
      \hline
      1  & 86.7 & 86.5 & 86.6 &  84.5  &  86.4 & 85.4 & 78.2 & 78.2 & 78.2 \\
      2 & 86.5 & 86.0 & 86.3 & 84.1 & 86.9 & 85.1 & 76.6 & 79.9 & 78.2 \\
      3 & 86.2 & 86.9 & 86.6 & 83.0 & 87.4 & 85.1 & 77.7 & 78.1 & 77.9  \\
	  4 & 85.6  & 86.6 & 86.1  & 83.1 & 86.8 & 84.9 & 75.2 & 80.2 & 77.6 \\
	  5 &84.8  & 88.1  & 86.4 &  83.7 & 86.9 & 85.3 & - & - & - \\
	  6 &  84.9 & 87.4 & 86.1 & 82.6 & 87.8 & 85.1  & - &  - & - \\
      label size & 84.3 & 87.4 & 85.8  & 82.6 & 87.5 & 85.0 & 76.7 & 77.6 & 77.2 \\
      2 * label size &  84.2 & 87.5 & 85.8 & 82.6 & 87.0 & 84.7 & 77.0 & 76.7 & 76.9  \\
	  3 * label size &  83.5 & 87.4 & 85.4 & 82.3 & 86.5 & 84.3 & 76.6 & 77.3 & 77.0  \\
      \bottomrule
    \end{tabular}		
  \end{center}
  \caption{\label{tab:latent_labels} Results of varying the number of latent labels $|\mathcal{L}_l|$ on the ACE2004, ACE2005, and GENIA dataset. The observed label size is 7 in ACE2004 and ACE2005 datasets, and 5 in GENIA dataset. As the size of the latent labels increases, we observe an interesting trend of increasing recall and decreasing precision. This means that the model is making more radical predictions, targeting on coverage (recall) rather than confidence (precision). See the corresponding paragraph for detailed discussions. 
  } 
  
\end{table*}

\begin{table*}[t!]
  \begin{center}
    \begin{tabular}{@{} l c  @{}}
      \toprule
      Method & UAS \\
      \hline
      Random          & 10.83  \\ 
      Left-branching  & 5.61  \\ 
      Right-branching & 0.15  \\ 
      PO-TreeCRF (ours)            & 16.95 \\ 
      \hline 
      Oracle     & 78.51 \\ 
      \bottomrule 
    \end{tabular}		
  \end{center}
  \caption{ \label{exp:parsing_eval} Alignment to syntactical constituency trees. 
  Oracle trees are obtained by using CoreNLP to parse the sentences, then left-binarized to CNF. 
  All the reported methods align poorly to the syntactical constituency trees, 
  however, our method still gives higher UUAS than all baseline, 
  meaning that a small fraction of tree structures predicted by our model are aligned with the syntactical structures.} 
  
\end{table*}

\begin{table*}[t!]
	\begin{center}
		\small
		\begin{tabular}{@{} l c c  @{}}
			\toprule 
			Method & \textsc{Masked Inside} & Vanilla Partial Marginalization \\
			\hline
			GPU Time & 3m20s & 8m10s \\
			CPU Time & 24m & 26m \\
			\bottomrule
		\end{tabular}		
	\end{center}
	\caption{\label{tab:inf_speed} Time for training one epoch on ACE2004. 
		GPU Nvidia P100, CPU Intel 2.6Hz quad-core i7. The vanilla partial marginalization algorithm does not support sentence batchification since different sentences have different tree structures, thus being significantly slower than the masked inside.} 
\end{table*}





\subsection{Detail Results of 5 Runs}

Table~\ref{tab:full_result} gives the full results of all 5 runs. 
We report the results simply to check if the model is sensitive to random seeds. 
The numbers show that the F1 scores do not vary much with different initialization, which means that the model is consistent under random initialization. 

\subsection{Number of Latent Labels}
Table~\ref{tab:latent_labels} shows the performance change as we increase the number of latent states. 
When the number of latent states is relatively small (less than 5), the performance does not vary much. 
So we report the performance for 1 latent state in the main paper. 
We also note that when we increase the number of latent states to be larger than the number of observed states, the performance drops significantly.

Table~\ref{tab:latent_labels} shows an interesting trend of changing the size of latent labels: as we increase the number of latent labels, the precision is gradually decreasing while the recall is gradually increasing. 
To explain this observation, we make the following conjecture: 
(a) the sum of the prior probability of observed labels and latent labels is 1, and the prior probability of observed labels is fixed in the training set, so different latent labels would share the rest prior probability; 
(b) as the number of latent labels increases, each latent label would in average receive less probability;
(c) during CKY decoding, since the probability of each latent label decreases on average, it is more likely that a tree involving more observed labels would be of higher probability, resulting the algorithm to decode out trees with more observed labels;
(d) consequently, the model makes more radical predictions, leading to higher coverage of entities (higher recall) but lower precision. 
We note this reasoning chain is only a rough sketch, and we have not yet able to investigate the detailed mechanism. 
We leave it to future work. 

\subsection{Alignment to Syntax Trees}

We further compare the trees inferred by our model with the syntactical constituency trees of the same sentence.
We use unlabeled attachment score (UAS) as the similarity metric.
We report the similarity on the training set.
Specifically, we use Stanford CoreNLP to obtain the  oracle constituency trees of the training sentences, binarize these trees to CNF using an always-left strategy. 
The UAS of the oracle trees means the similarity between the left-binarized oracle trees to the original oracle trees. 
We compare: (a) random trees (b) left-branching trees (d) right-branching trees (d) trees inferred from our NER model to the oracle trees. 
All baseline trees align poorly to the oracle constituency trees, while ours have the highest UAS. 
We emphasize that our trees are trained under a NER objective, and do not necessarily need to be similar to syntax trees. 
However, there is still a  fraction of the tree structures inferred from our model that is similar to the syntactical constituency structures. 
We leave further investigation about the relationship between our latent trees and syntactical constituency trees to future work.

\subsection{Batchification Speed with \textsc{Masked Inside}}
We note that the vanilla partial marginalization algorithm (Algorithm 2) in the main paper does not support sentence-level bachification because different sentences have different tree structures. 
This inefficiency is addressed by the \textsc{Masked Inside} algorithm.
To see how sentence-level batchification leads to further speed improvement, we report the speed comparison in Table~\ref{tab:inf_speed}. 
Generally, \textsc{Masked Inside} is significantly faster than a vanilla partial marginalization algorithm.

\end{document}